\documentclass{article} 

\usepackage{microtype}
\usepackage{graphicx}
\usepackage{algorithmic}
\usepackage{algorithm}
\usepackage{caption}
\usepackage{natbib}
\usepackage{subcaption}
\usepackage{amsmath}
\usepackage{amssymb}
\usepackage{amsfonts}
\usepackage{booktabs} 
\usepackage{hyperref}

\usepackage[accepted]{icml2020}
\usepackage{chngcntr}

\usepackage{bm}
\usepackage{import}
\usepackage{xspace}

\usepackage{hyperref}
\usepackage{url}
\newcommand{\deriv}[2][]{\frac{\partial#1}{\partial#2}}
\usepackage{graphicx}
\usepackage{subcaption}
\long\def\/*#1*/{}
\newcommand{\pvnfull}[0]{Policy Evaluation Network}
\newcommand{\pvnfulls}[0]{\pvnfull s}
\newcommand*{\defeq}{\mathrel{\vcenter{\baselineskip0.5ex \lineskiplimit0pt
                     \hbox{\scriptsize.}\hbox{\scriptsize.}}}%
                     =}
\def\pvnfun{{\bm \psi}}
\def\bphi{{\bm \phi}}

\def\btheta{{\bm \theta}}
\def\pvnweights{{\bm w}}
\def\pvn{PVN\xspace}
\def\pvns{PVNs\xspace}
\def\probe{\bphi}

\def\nbins{m}
\def\nbatch{n}
\def\ndimstates{k}

\icmltitlerunning{\pvnfulls}

\begin{document}

\twocolumn[
\icmltitle{\pvnfulls}
\begin{icmlauthorlist}
\icmlauthor{Jean Harb}{mcgill,dm}
\icmlauthor{Tom Schaul}{dm}
\icmlauthor{Doina Precup}{dm}
\icmlauthor{Pierre-Luc Bacon}{udem}
\end{icmlauthorlist}

\icmlaffiliation{mcgill}{Mila - McGill University}
\icmlaffiliation{dm}{DeepMind}
\icmlaffiliation{udem}{Mila - University of Montreal}
\icmlcorrespondingauthor{Jean Harb}{jean.merheb-harb@mail.mcgill.ca}

\icmlkeywords{Machine Learning, Reinforcement Learning}

\vskip 0.3in
]
\printAffiliationsAndNotice{}

\begin{abstract}
Many reinforcement learning algorithms use value functions to guide the search for better policies. These methods estimate the value of a single policy while generalizing across many states. The core idea of this paper is to flip this convention and estimate the value of many policies, for a single set of states. This approach opens up the possibility of performing direct gradient ascent in policy space without seeing any new data. The main challenge for this approach is finding a way to represent complex policies that facilitates learning and generalization. To address this problem, we introduce a scalable, differentiable fingerprinting mechanism that retains essential policy information in a concise embedding. Our empirical results demonstrate that combining these three elements (learned \pvnfull, policy fingerprints, gradient ascent) can produce policies that outperform those that generated the training data, in zero-shot manner.
\end{abstract}

\section{Introduction}
Value functions are core quantities estimated by most reinforcement learning (RL) algorithms, and used to inform the search for good policies. Usually, value functions receive as input a description of the state (or state-action pair) and estimate the expected return for some distribution of inputs, conditioned on some behavior, or policy.  When the policy changes, value estimates have to keep up. 

One way to view this process is that value functions are trained using large amounts of transitions, coming from the set of policies that have been used by the agent in the past, without seeing from which policy each transition came. As the policy changes, the value function estimate is still influenced by previously seen policies, possibly in an unpredictable way, because the policy is not typically represented as an input. Our goal is to explore the idea of pushing the agent to generalize its value representation among different policies, by providing a policy description as input. We hypothesize that an agent trained in this way  could predict the value of a new, unseen policy with sufficient accuracy to guide further search through policy space. In particular, if we were to train a network to estimate the value of a policy from an initial start state distribution, we would have access to the gradient of the expected return with respect to the policy's parameters and could use it to improve the policy. The value function would not need to take states as input at all, relying instead on an encoding of the policy itself in order to predict the expected return of the entire episode, from the starting state distribution.

Our first contribution is to introduce the \pvnfull\ (\pvn), a network that learns to predict the expected return of a policy in a fully differentiable way. Using a dataset of policy networks along with their returns, the \pvn\ is trained with supervised learning.

However, it is not trivial to embed a policy in a way that allows the embedding to be sufficiently informative for the value function, yet not too large. For example, the naive approach of flattening a policy network into a large vector loses information on the dependencies between layers, while the vector can still become intractably large. The second major contribution of this paper is a new method to fingerprint a policy network in order to create an embedding that is sufficiently small, yet retains information about the network structure.

Finally, we introduce a novel policy gradient algorithm, which performs gradient ascent through a learned \pvn in order to obtain, in zero-shot manner, policies that are superior to any of those evaluated during training.

We present small scale experiments to illustrate the behavior  of our approach, then present experiments that show how fingerprinting allows us to evaluate not only linear policy networks, but also  multi-layered ones.

\section{Background}
In RL, an agent's goal is to learn how to maximize its return by interacting with an environment modelled, which is usually assumed to be a Markov Decision Processes (MDP) $\langle \mathcal{S}, \mathcal{A}, \mathcal{P}, \gamma, r \rangle$, 
where $\mathcal{S}$ is a set of states, $\mathcal{A}$ is a set of actions, $\mathcal{P}: \mathcal{S} \times \mathcal{A} \times \mathcal{S} \rightarrow [0,1]$ represents the environment dynamics, $\gamma \in [0,1]$ is a discount factor and 
$r: \mathcal{S} \times \mathcal{A} \rightarrow \mathbb{R}$
is the reward function. A randomized policy $\pi: \mathcal{S} \times \mathcal{A} \rightarrow [0,1]$ is a distribution over actions, conditioned on states.

In this paper, we consider the problem of finding the parameters $\theta \in \mathbb{R}^k$ of a stochastic policy $\pi_\theta$ which maximize the expected discounted return from a distribution over initial states $d_0$: $J(\theta) \defeq \mathbb{E}\left[\sum_{t=0}^\infty \gamma^{t} r(S_t, A_t)\right|S_0 \sim d_0]$. 
The policy gradient theorem \citep{sutton2000policy} shows that gradient of this objective is given by:
\begin{equation}
    \deriv[J(\theta)]{\theta} = \sum_{s \in \mathcal{S}} d_{\pi_\theta}(s)\sum_{a \in \mathcal{A}} \deriv[\pi_\theta(a|s)]{\theta} Q_{\pi_\theta}(s,a) \enspace \label{eq:grad},
\end{equation}
and where $Q_{\pi_\theta}(s,a)$ is the action-value function corresponding to policy $\pi_\theta$ and $d_{\pi_\theta}$ is a discounted weighting of states:
\begin{align*}
d_{\pi_\theta}(s) \defeq \sum_{s_0 \in \mathcal{S}} d_0(s_0) \sum_{t=0}^\infty \gamma^t P_{\pi_\theta}(S_{t} = s | S_0 = s_0) \enspace .
\end{align*}
Here, $P_{\pi_\theta}(S_{t} = s | S_0 = s_0)$ is the probability of reaching $s$ from 
$s_0$ at time step $t$, given the environment dynamics and the fact that actions are chosen from $\pi_\theta$.

The gradient (\ref{eq:grad}) is typically estimated with samples taken under the undiscounted distribution induced by $\pi_\theta$ in the MDP~\citep{thomas2014}. In actor-critic architectures~\citep{sutton1984}, a learned estimate of the action-value function $Q_{\pi_\theta}$ is usually maintained in a two-timescale manner~\citep{konda1999}, ie., by allowing the iterates of $Q_{\pi_\theta}$ to converge faster than  the policy parameters. This requirement is crucial for the stability of such methods in practice~\citep{Fujimoto2018}.

In this paper, we propose a new gradient-based optimization method which does not rely on maintaining a value function estimate in this two-timescale fashion. Our key observation is that rather than using a stochastic gradient of the performance measure, we can learn an estimate of $J(\theta)$ directly (as a neural network) and compute a deterministic gradient from it using any automatic differentiation package. We will now introduce the core idea of this approach.

\section{\pvnfulls}
\label{sec:penfull}
By definition, a value function represents the expected return associated with a given policy. So, one could expect, as in regression methods, that training could be done on some policies and generalize to others. But, value function approximation methods are not typically designed with the goal of leveraging any information about the policy itself in order to  generalize to new policies. Conceptually, though, there exists a function capable of computing the expected return for any policy in a \textit{zero-shot} fashion: the performance measure itself. In vector notation, we can write the performance measure explicitly as:
\begin{equation}
    J({\bm \theta}) \defeq {\bm d_0}^\top ( \mathbf{I} - \gamma \mathbf{P}_{\pi_\theta})^{-1}\mathbf{r_\theta}_\pi \enspace . \label{eq:explicit_perf}
\end{equation}
where $\mathbf{P}_{\pi_\theta}$ and $\mathbf{r_\theta}_\pi$ are the transition matrix and reward model induced by $\pi_\theta$. Hence, there is a function of $\theta$ that can compute the value of $\pi_\theta$.  

\pvnfulls\ aim to approximate this function $J$, so that the gradient of a parameterized policy can be obtained instantaneously. Furthermore, we show that this can be achieved in a completely model-free fashion, without having to estimate $\mathbf{P}_{\pi_\theta}$ and $\mathbf{r}_{\pi_\theta}$, or to form any of the matrices in \eqref{eq:explicit_perf}.

\textbf{Distributional Predictions.} While approximating $J$ could be directly cast as a regression problem, we  view it instead as a classification problem: that of predicting the \textit{bucket} index corresponding to the estimated return of a given input policy. This strategy has proven to be effective in the training of deep neural networks,  both in the supervised learning regime~\citep{Oord2016} and in reinforcement learning~\citep{Bellemare2017}. Predicting buckets instead of exact real values provides a regularization effect,  similar to the idea of learning with auxiliary tasks~\citep{Caruana1997,Sutton2011,Jaderberg2017,lyle2019comparative}.

Inspired by \citet{bellemare2017distributional}, we discretize the set of sampled returns into buckets of the same size. The \pvn\ then outputs a probability distribution over the set of buckets, and the loss we use is the KL-divergence between the predicted and target distributions. In practice, the target is determined by rolling out multiple episodes per policy and discretizing the resulting returns. 

\textbf{Maximizing Undiscounted Returns.} 
RL algorithms have traditionally been viewed as optimizing for the discounted return, with some given discount factor $\gamma$. However, the practice of policy gradient methods has evolved towards the use of discounting  as a knob to control the bias-variance tradeoff~\citep{Cao1998,Baxter2001}. Controlling the variance of the gradient estimator is of paramount importance, because it can grow exponentially in the horizon~\citep{Glynn2019}. Hence, rather than seeing the discount factor as being part of the task specification, practitioners  tend to view it as a hyperparameter  to be optimized~\citep{schulman2016}. 

We are ultimately interested in the undiscounted performance of a policy as in \citet{thomas2014}. Because  \pvnfulls\ form a deterministic gradient of an approximated loss, they  sidestep the need to pick a discount factor and optimize a proxy objective. Our experiments show that our approach is capable of optimizing for the undiscounted performance directly, and outperforms the state-of-the art discounted methods.

\textbf{Learning Objective for the \pvn.}
When the \pvn (denoted $\pvnfun$) provides a categorical output over return bins, we need to specify how a scalar  expected return estimate is obtained from this output. A straightforward approach consists in using the midpoint of each interval as a representative of the return, which we then weight by the predicted probability  (denoted by $\pvnfun_i(\btheta)$) for this interval:
\begin{align*}
    h &\defeq \frac{1}{\nbins}\left(G_\text{max} - G_\text{min} \right) \\
    \hat J({\bm \theta}) &\defeq \sum_{i=0}^{\nbins-1} \pvnfun_i(\btheta) \left( G_\text{min} + \frac{h}{2} + ih\right) \enspace ,
\end{align*}
where $\nbins$ is the number of bins, $h$ is the width of each bin, and $G_\text{min}$ and $G_\text{max}$ are the minimum and maximum returns observed in the dataset. However, rather than minimizing the L2 distance between $\hat{J}(\btheta)$ and samples of $J(\btheta)$, we use a classification loss: the KL-divergence between the output of the \pvn and an empirical probability mass function over the discretized returns. We then view a \pvn as a function of the form $\pvnfun(\btheta;\pvnweights)$ where $\pvnweights$ are the parameters of the network itself and $\btheta$ are the policy network parameters fed as input. We then use stochastic gradient descent to minimize the expected KL loss: 
\begin{align*}
    \min_{\pvnweights} \mathbb{E}\left[D_\text{KL}( \hat{P}_\btheta \,||\, \pvnfun(\btheta; \pvnweights)) \right]\enspace ,
\end{align*}
where the expectation is taken under a given distribution over policy network weights (random in our experiments) and $\hat{P}_\btheta$ is obtained from a histogram of the returns induced by a policy $\pi_\theta$. 

\section{Network Fingerprinting}
\label{sec:probe}

While the performance measure $J$ is by definition a function of the policy network parameters, we need to figure out how to provide the policy as an input to a \pvn, in a way which does not require too much space and also leads to good generalization among policies. If we try to naively flatten the policy into a vector input, we run into severla issues. First, the number of weights in the policy network can be very large, which requires a weight matrix in the input layer of the \pvn of at least the same size. Second, the dependencies between layers contain valuable information, which is lost by a direct concatenation of the parameters. 

\begin{figure}[th]
    \centering
    \includegraphics[width=0.45\textwidth]{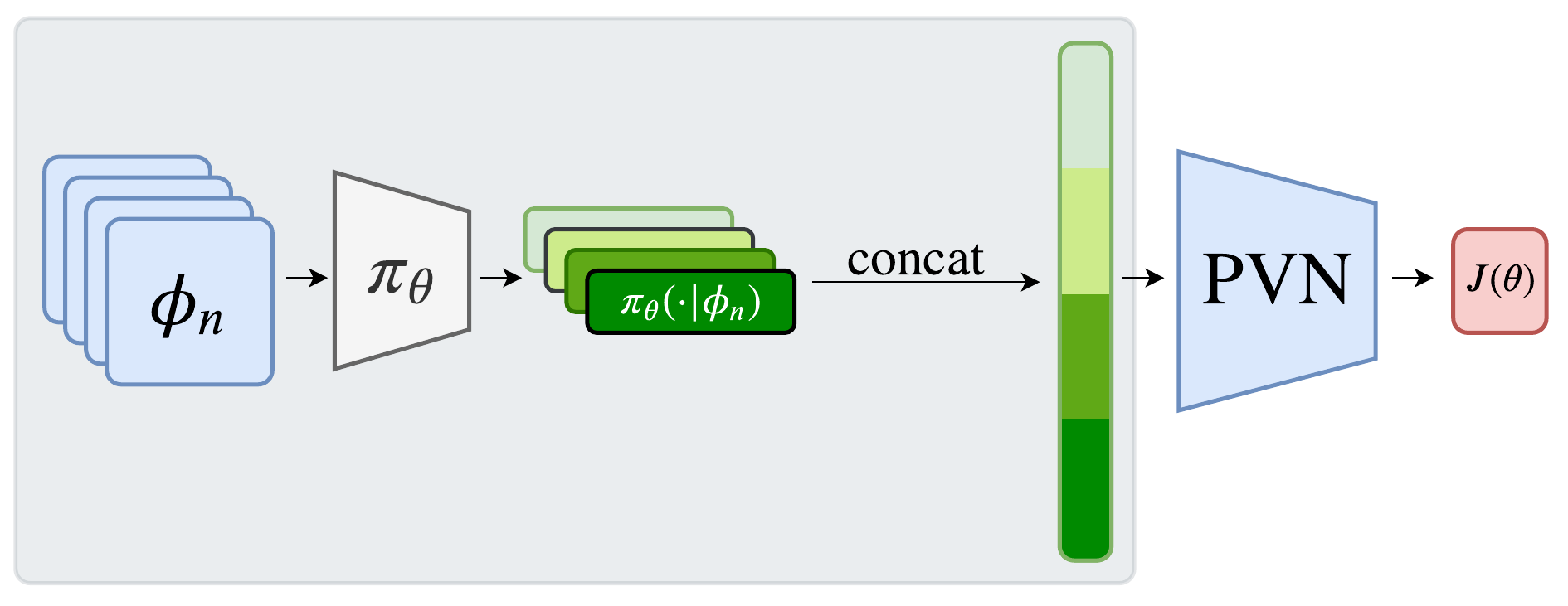}
    \caption{Diagram of the complete \pvnfull\ setup, including Network Fingerprinting (in gray). The blue color of the probing states and \pvn\ indicates that they can be seen as one set of weights, trained in unison.}
    \label{fig:full_pen_arch}
\end{figure}

To address these issues, we propose \textit{network fingerprinting}: a methodology which allows us to learn a neural network embedding in a fully differentiable way, independently of the number of parameters in the policy network. The intuitive idea is to characterize the response of a policy network by feeding it a set of $\nbatch$ learned \textit{probing states} as input $\probe  = [\mathbf{s}_1, \ldots, \mathbf{s}_\nbatch] \in \mathbb{R}^{\nbatch \times \ndimstates}$, 
where $\ndimstates$ is the dimensionality of a state vector. $\probe$ is a synthetic input to the policy network whose sole purpose is to elicit a representative output. This output can be a distribution over discrete actions or continuous action vectors, which we then concatenate and use as input to the \pvn. In discrete action spaces, we can view a \pvn as function $\pvnfun: \mathbb{R}^{n|\mathcal{A}|} \to \mathbb{R}^\nbins$ receiving a $|\mathcal{A}|n$-dimensional vector of probabilities over actions and returning a distribution over $\nbins$ categories corresponding to the discretized return. The output of the \pvn is then computed through the composition $\pvnfun(\pi_\theta\left(\cdot \,\middle|\, \probe\right))$

\textbf{Motivation for Network Fingerprinting.} To see why this might be a viable method, we can show an equivalence between using $\nbatch$ probing states and having $\nbatch$ hidden units in the first layer of a \pvn which evaluates  linear policies. Consider a linear, deterministic policy in a 1D action space, defined by $\pi_{\theta}(\mathbf{s}) = \theta^{\top}\mathbf{s} \in \mathbb{R}$.
Probing this policy in $\nbatch$ states $\probe$ produces the fingerprint vector $\mathbf{y} = \pi_{\theta}(\probe) = \theta^{\top}\probe \in \mathbb{R}^{\nbatch}$.
On the other hand, feeding the full policy weights $\theta$ as input to a \pvn\ with $\nbatch$ hidden units in the first layer will produce the hidden activations $\mathbf{y} =\theta^{\top}\mathbf{W} \in \mathbb{R}^{\nbatch}$ at that layer. Clearly then, there is a choice of probing states $\probe$ and network initialisation $\mathbf{W}$ that produces exactly equivalent results.

\textbf{Choosing Probing States.} The choice of probing states is important. One possible approach is to randomly generate states. Since we know the dimensionality of inputs to the policy, one could simply choose a sampling distribution and create fake states made of noise. Alternatively, we could also sample states from the environment, because we are interested in learning to evaluate the performance of the policy on states that we actually observe.

Because the entire \pvn architecture is differentiable, we can also use backpropagation to learn the probing states, as we can see in Fig.~\ref{fig:full_pen_arch}. These states can be viewed as weights of the \pvn, helping extract information from the policy to improve prediction accuracy. In this paper, we adopt an hybrid approach. First, we initialize the probing states by sampling random noise. Then, we  refine those probing states throughout learning, while learning the \pvn weights jointly. When using the classification loss, our minimization problem becomes:
\begin{align*}
    \min_{\pvnweights, \probe} \mathbb{E}\left[D_\text{KL}( \hat{P}_\btheta \,||\, \pvnfun(\pi_\btheta(\cdot | \probe); \pvnweights)) \right]\enspace ,
\end{align*}
which we optimize by stochastic gradient descent. 

\section{Policy Improvement By Gradient Ascent}

Using a trained \pvn, it is possible to do gradient ascent in the space of parameterized policies without having to interact with the environment. Because the \pvn is an approximation to the real performance measure in functional form, we can directly apply automatic differentiation to obtain an exact deterministic gradient. For example, when using network fingerprinting inside a \pvn $\pvnfun$ parameterized by $\pvnweights$, our gradient ascent procedure computes the iterates:
\begin{align*}
    \btheta_{t+1} = \btheta_t + \eta_t \nabla_{\btheta_t} \pvnfun(\pi_{\btheta_t}(\cdot|\probe) ;\pvnweights)\enspace ,
\end{align*}
where $\eta_t$ is the learning rate at time $t$ and $\probe$ are the learned probe states.

A benefit of not having to interact with the environment to get fresh gradient estimates for every new policy is that we can do gradient ascent in parallel via our learned \pvn. This feature is particularly useful to escape local maxima, as many concurrent solutions can be maintained simultaneously as in \cite{jaderberg2017population}. Statistical active learning techniques \citep{Cohn1995} could make this process more efficient, but would require \pvns with epistemic uncertainty.

Because the gradient ascent procedure may lead us to regions of the policy space where fewer samples have been obtained to train the \pvn, being able to maintain multiple candidate solutions in parallel is particularly useful. To further avoid falling too quickly into the out-of-distribution regime, we can also limit the number of gradient steps and periodically verify that the performance is still increasing.

\section{Experiments}
In order to test our proposed method, we first create a set of policies by choosing a neural network architecture and initializing it a number of times. Second, we obtain the expected returns of these policies by averaging returns from a number of Monte-Carlo rollouts. This results in a dataset, where the policies are the inputs and the expected returns are the targets. Finally, we create a \pvn\ and regress on the dataset.

We ran experiments on a variety of environments of different complexities. We start with a simple 2-state MDP with known dynamics, which allows us to visualize how our algorithm works. We then move on to function approximation on the Cart Pole environment, where we analyze the effects of using Network Fingerprinting on single and multi-layered networks. Finally, we move to the Swimmer environment to show our algorithm's performance on a continuous control task.

\begin{algorithm}[tb]
   \caption{Train PVN with fingerprinting}
   \label{alg:pen2}
\begin{algorithmic}

    \STATE Initialize \pvn\ parameters $\pvnweights$, $\probe$, choose learning rate $\alpha$
    \FOR{step $s=1$ {\bfseries to} $S$}
    \STATE Sample training batch $\{(\pi_{\theta_i}, R_i)\}_{i=1}^B \sim \mathcal{D}$
    \STATE $\mathcal{L}(\pvnweights, \probe)\leftarrow D_{\text{KL}}( \text{hist}(R_{B}) || \pvnfun({\bf \pi}_{\theta_{B}}(\cdot | \probe); \pvnweights ))$
  \STATE update parameters $\pvnweights \leftarrow \pvnweights - \text{adam}(\alpha, \nabla_\pvnweights \mathcal{L}(\pvnweights, \probe))$\;
  \STATE update parameters $\probe \leftarrow \probe - \text{adam}(\alpha, \nabla_\probe \mathcal{L}(\pvnweights,\probe))$\;
  \ENDFOR
\end{algorithmic}
\end{algorithm}
  
\begin{algorithm}[tb]
   \caption{Gradient Ascent Through a Trained \pvn}
   \label{alg:pen3}
\begin{algorithmic}
    \STATE $\Theta \leftarrow \varnothing$
    \FOR{ascent policies $i=1$ {\bfseries to} $A$}
    \STATE Initialize policy $\pi_i$ parameters $\theta_i$ using Glorot Initialization
    \STATE $\mathcal{D} \leftarrow \varnothing$
    \FOR{ascent steps $j=1$ {\bfseries to} $T$}
    \STATE $\theta_i \leftarrow \theta_i + \beta \nabla_{\theta_i} \hat J(\theta_i)$
    \STATE Sample Monte-Carlo return $G_{MC}$ using policy $\pi_{\theta_i}$
    \STATE $\mathcal{D} \leftarrow \mathcal{D} \cup (\theta_i, G_{MC})$
    \ENDFOR
    \STATE $\Theta \leftarrow \Theta \cup (\text{argmax}_{\theta_i}(\mathcal{D}), \text{max}(\mathcal{D}))$
    \ENDFOR
    \STATE \textbf{return} $ \text{argmax}_{\theta}(\Theta)$
\end{algorithmic}
\end{algorithm}

\subsection{Polytope}
\label{sec:polytope}

\begin{figure*}[t]
\centering
\begin{subfigure}[t]{0.31\textwidth}
  \includegraphics[width=.95\linewidth]{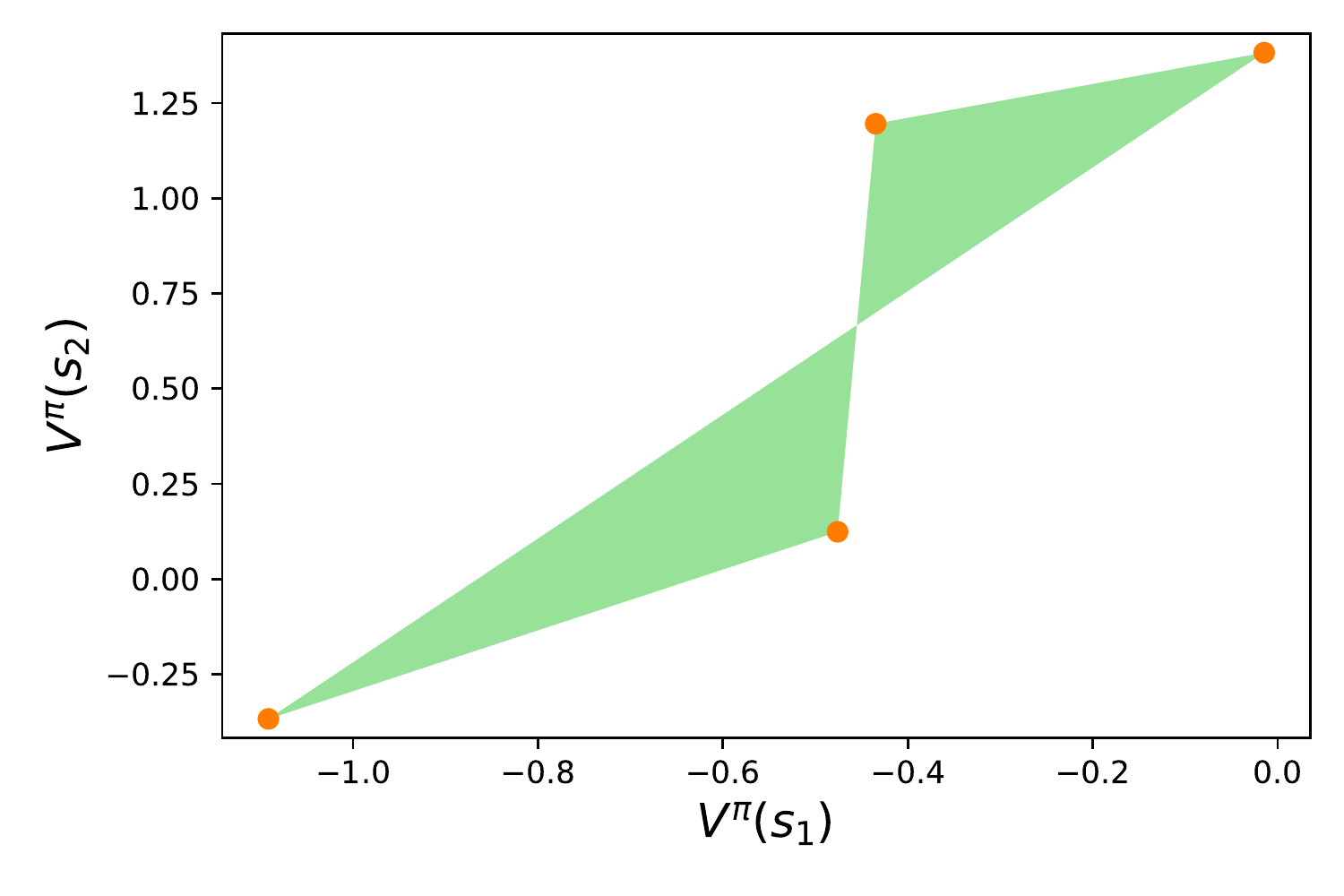}
    \caption{This polytope represents the space of policies (in green) in value space. The corners (in orange) are deterministic policies.}
    \label{fig:example_polytope}
\end{subfigure}
\hfill
\begin{subfigure}[t]{.31\textwidth}
  \centering
  \includegraphics[width=.95\linewidth]{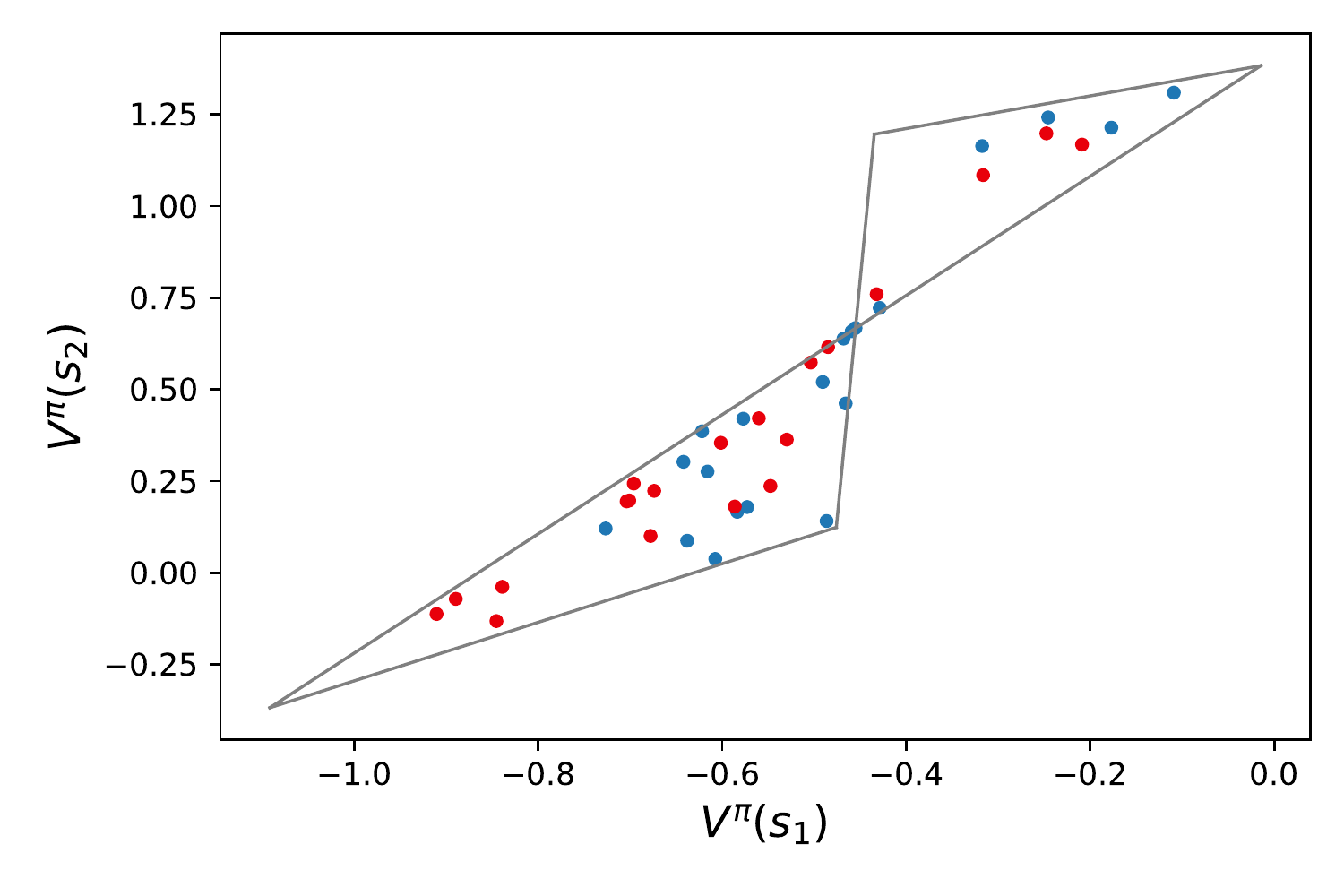}
  \caption{Sampled training set (in blue) and test set (in red) policies in the value polytope. }
  \label{fig:training_polytope}
\end{subfigure}%
\hfill
\begin{subfigure}[t]{.31\textwidth}
  \centering
  \includegraphics[width=.95\linewidth]{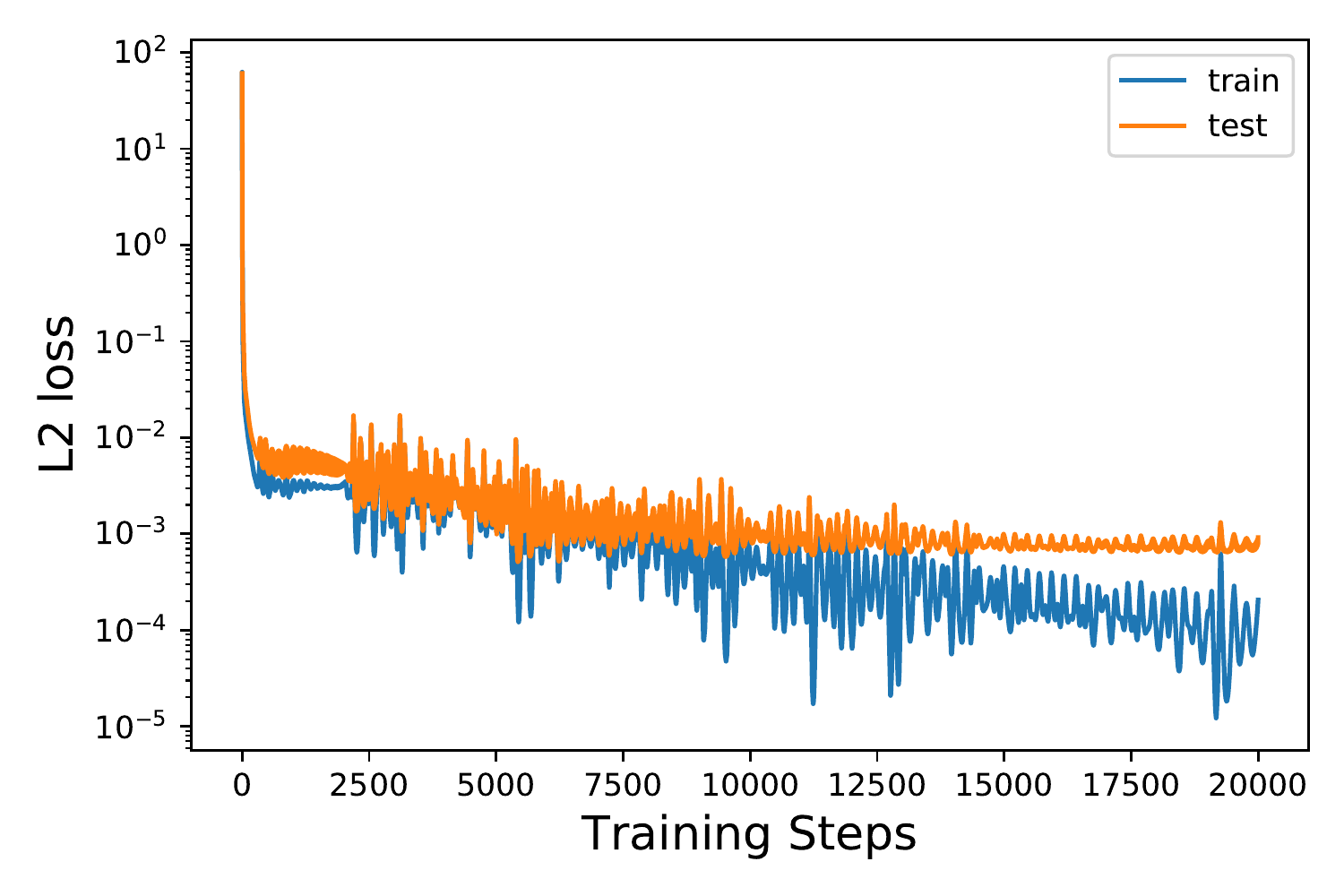}
  \caption{Training curves on training and test sets.}
  \label{fig:trained_polytope}
\end{subfigure}
\caption{Visualization of a value polytope and a sampled dataset of policies. Training curves show a \pvn can learn to generalize and predict the points in the test set.}
\label{fig:polytope_data}
\end{figure*}

\begin{figure*}[ht]
\begin{center}
\begin{subfigure}[t]{.32\textwidth}
  \centering
  \includegraphics[width=.95\linewidth]{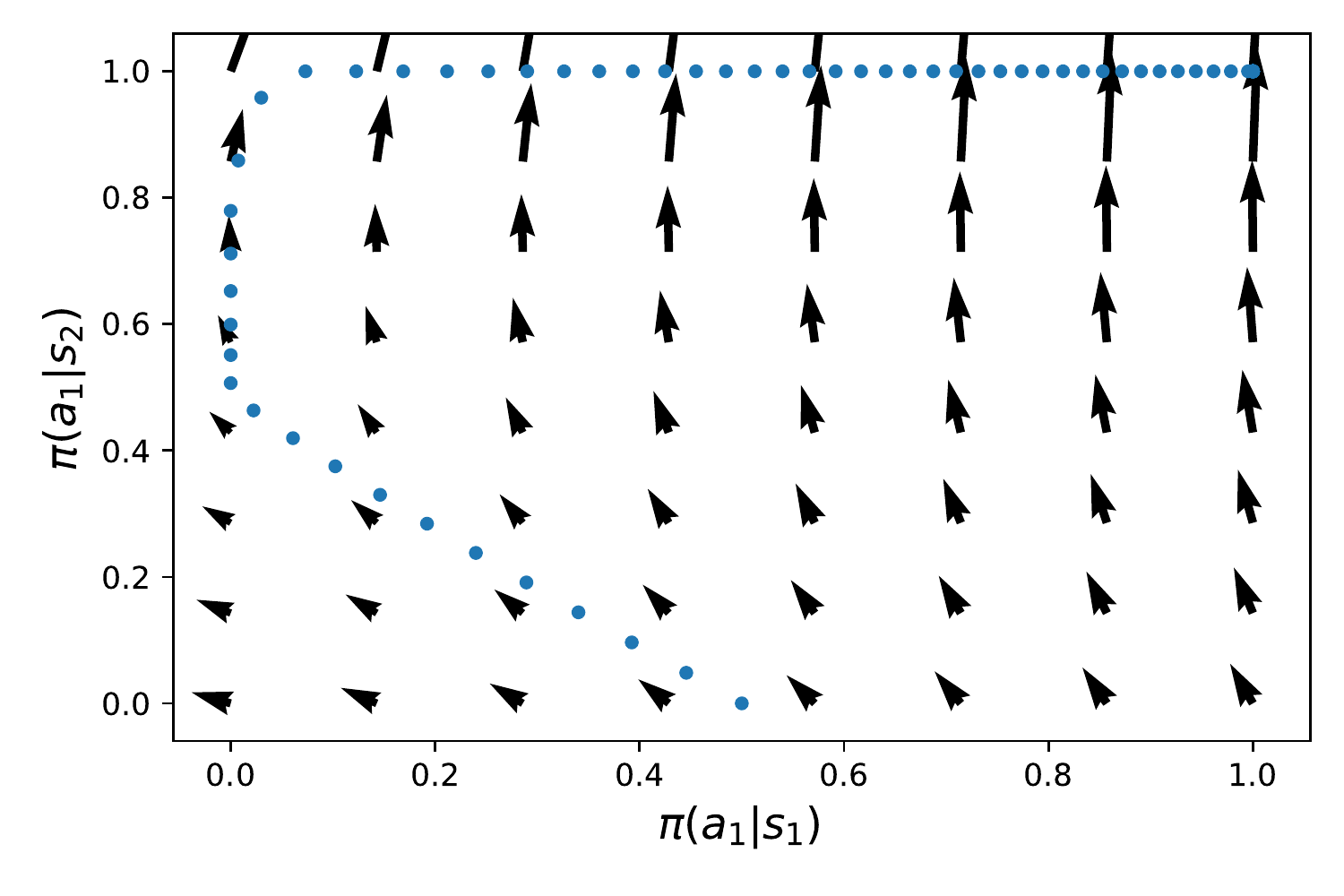}
  \caption{Exact gradient field of the value in policy space.}
  \label{fig:exact_grid}
\end{subfigure}%
\hfill
\begin{subfigure}[t]{.32\textwidth}
  \centering
  \includegraphics[width=.95\linewidth]{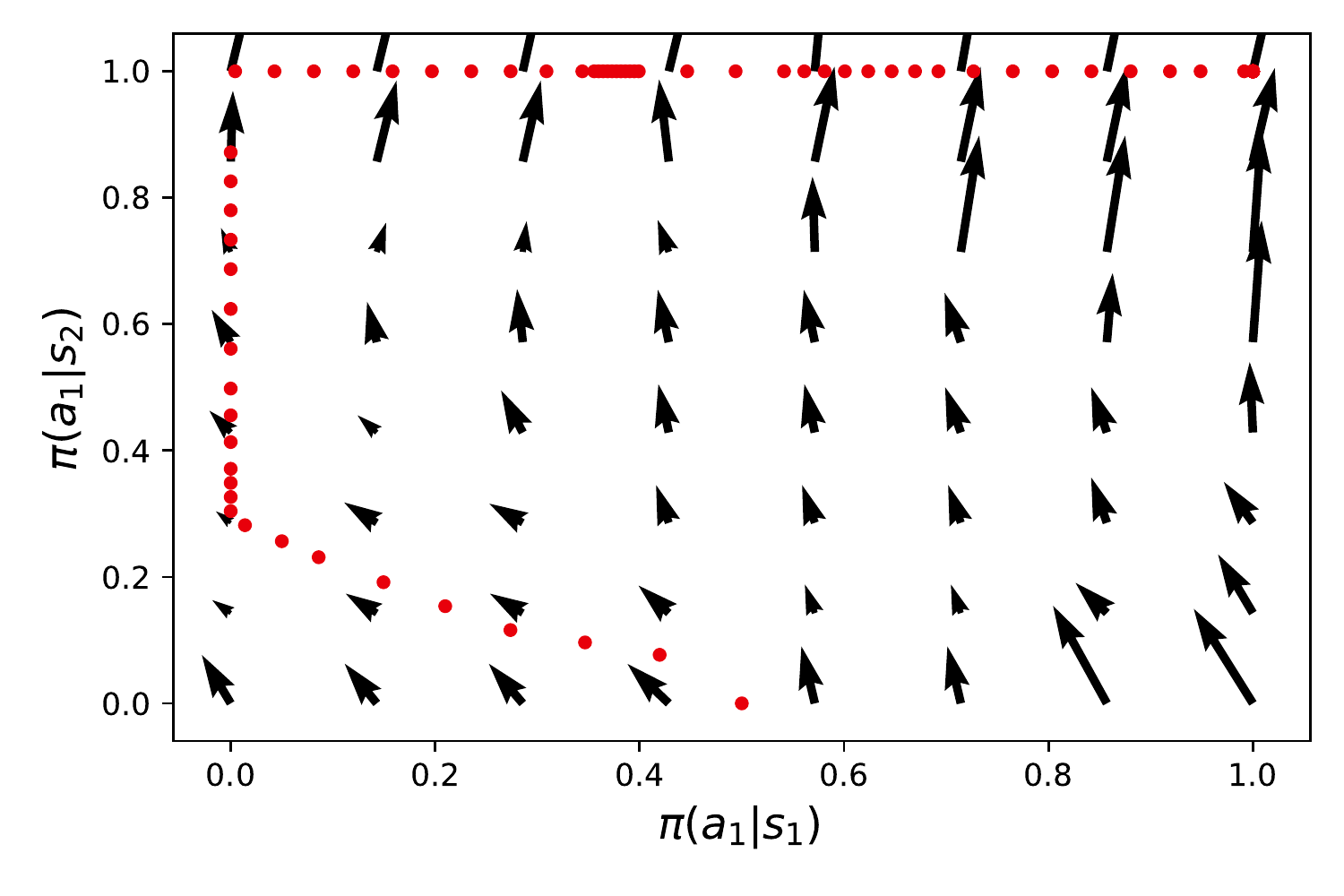}
  \caption{Approximated gradient field of the value in policy space, calculated from a trained \pvn.}
  \label{fig:learned_grid}
\end{subfigure}%
\hfill
\begin{subfigure}[t]{.32\textwidth}
  \centering
  \includegraphics[width=.95\linewidth]{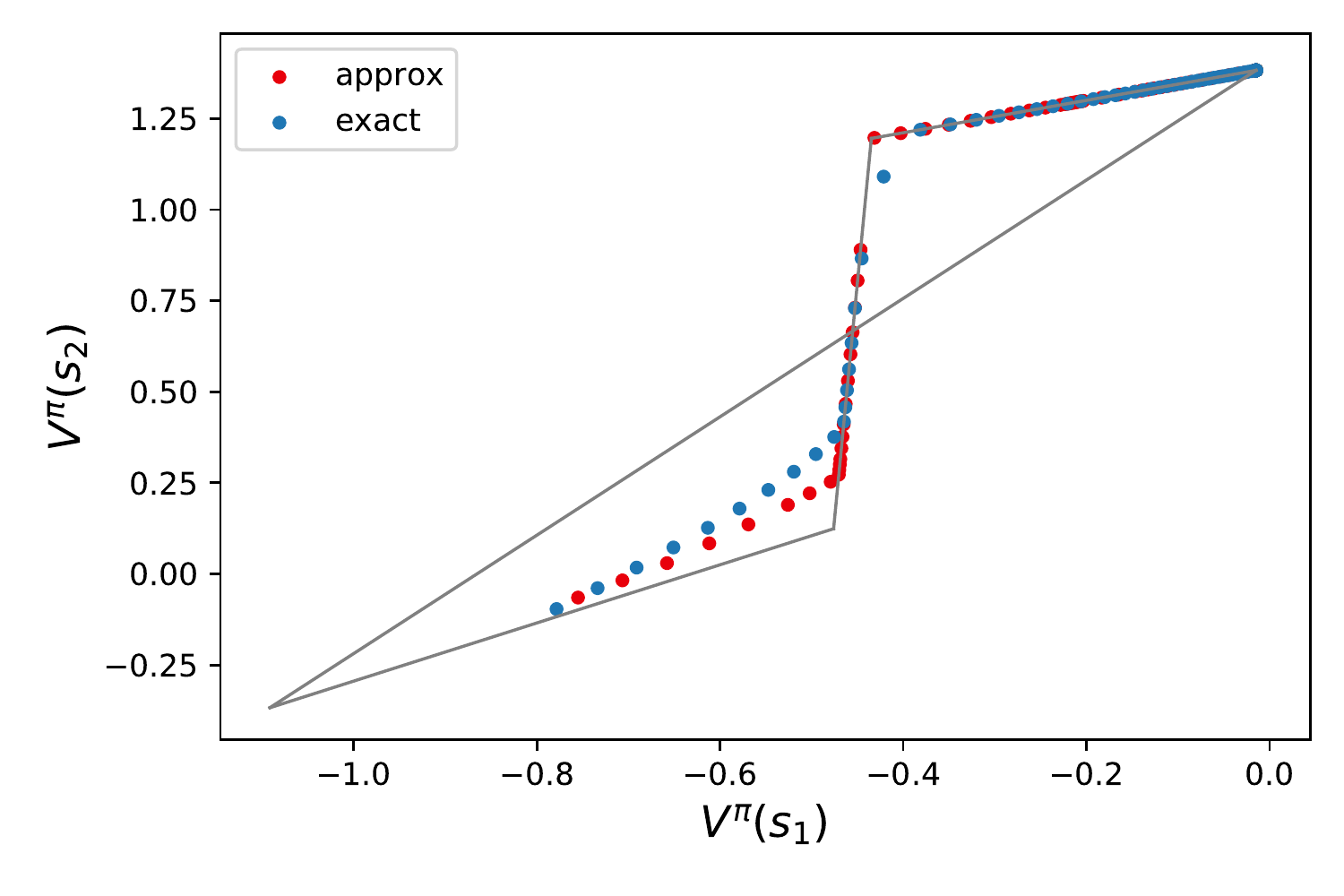}
  \caption{Comparison of gradient ascent through the exact and approximated value functions.}
  \label{fig:ascent_path}
\end{subfigure}
\end{center}
\caption{Comparison of gradient fields of the exact and approximated value functions. The two axes in Figures \ref{fig:exact_grid} and \ref{fig:learned_grid} are the policy spaces in each of the two states, and the arrows represent the gradient $\deriv[J(\theta)]{\theta}$ and $\deriv[\pvnfun(\btheta)]{\theta}$. The blue and red dots are steps of the gradient ascent process, mapped onto the polytope in Figure \ref{fig:ascent_path}. Both ascents were run for 100 steps.}
\label{fig:grad_ascent}
\end{figure*}

For the first set of experiments, we built a 2-state, 2-action MDP (details described in Appendix),
for which we know the transition matrix and reward function. Having this information allows us to calculate exact expected returns and policy gradients for any policy,  $\nabla_\theta J(\theta)$.
Moreover, because the MDP is so small, we can visualize the value polytope~\cite{dadashi2019value}, as seen in Figure \ref{fig:example_polytope}. A value polytope maps the space of policies to value space, meaning that any one point in the polytope is a policy and its coordinates are its values in states 1 and 2. Note that the corners of the polytope represent deterministic policies. In this environment, a policy can be represented as a simple 2-dimensional vector, $\pi = [P(a_1|s_1), P(a_1|s_2)]$, from which we can infer $P(a_2|s) = 1 - P(a_1|s)$.

To run our experiments, we first obtain a random set of 40 policies by sampling from a uniform distribution over the policy space. Then we evaluate them, using eq. (\ref{eq:explicit_perf}), to get their values, and split the data into a training and test set. The two sets of policies can be seen in Figure \ref{fig:training_polytope}. Finally, we train a 2-layered neural network on this regression task, where the network takes policy vectors as input and outputs their expected return.

\textbf{Policy Search.} Once we have a learned \pvn, we can perform gradient ascent through the network to search for the optimal policy. First, we can compare the exact policy gradients $\deriv[J(\theta)]{\theta}$ and those calculated from the learned \pvn\ $\deriv[\pvnfun(\btheta)]{\btheta}$. In Figures \ref{fig:exact_grid} and \ref{fig:learned_grid}, we show the difference in the discretized gradient fields of the true policy gradients and the learned ones. As we can see, the gradient fields are quite similar, with only a few differences  around the edges of the policy space.

Finally, to perform gradient ascent, we start with an arbitrary policy, in this case $[0.5, 0]$, calculate its gradient through the learned \pvn, take a small step in that direction and repeat the process with the new policy. In Figures \ref{fig:exact_grid} and \ref{fig:learned_grid}, we compare performing gradient ascent with the exact policy gradient and with our approximated one. The dots represent the paths taken by the gradient ascent process, and we can see that while the path given by the  learned function is noisier, they still both converge to the optimal solution. Figure \ref{fig:ascent_path} compares the two paths in  polytope space, to give a different perspective. Both ascents performed a series of 100 policy gradient steps.

These results indicate that \pvns\ can both approximate policy values and be used to calculate policy gradients in a practical manner. 

\begin{figure*}[t]
\begin{center}
\begin{subfigure}[t]{0.5\textwidth}
  \centering
  \includegraphics[scale=0.4]{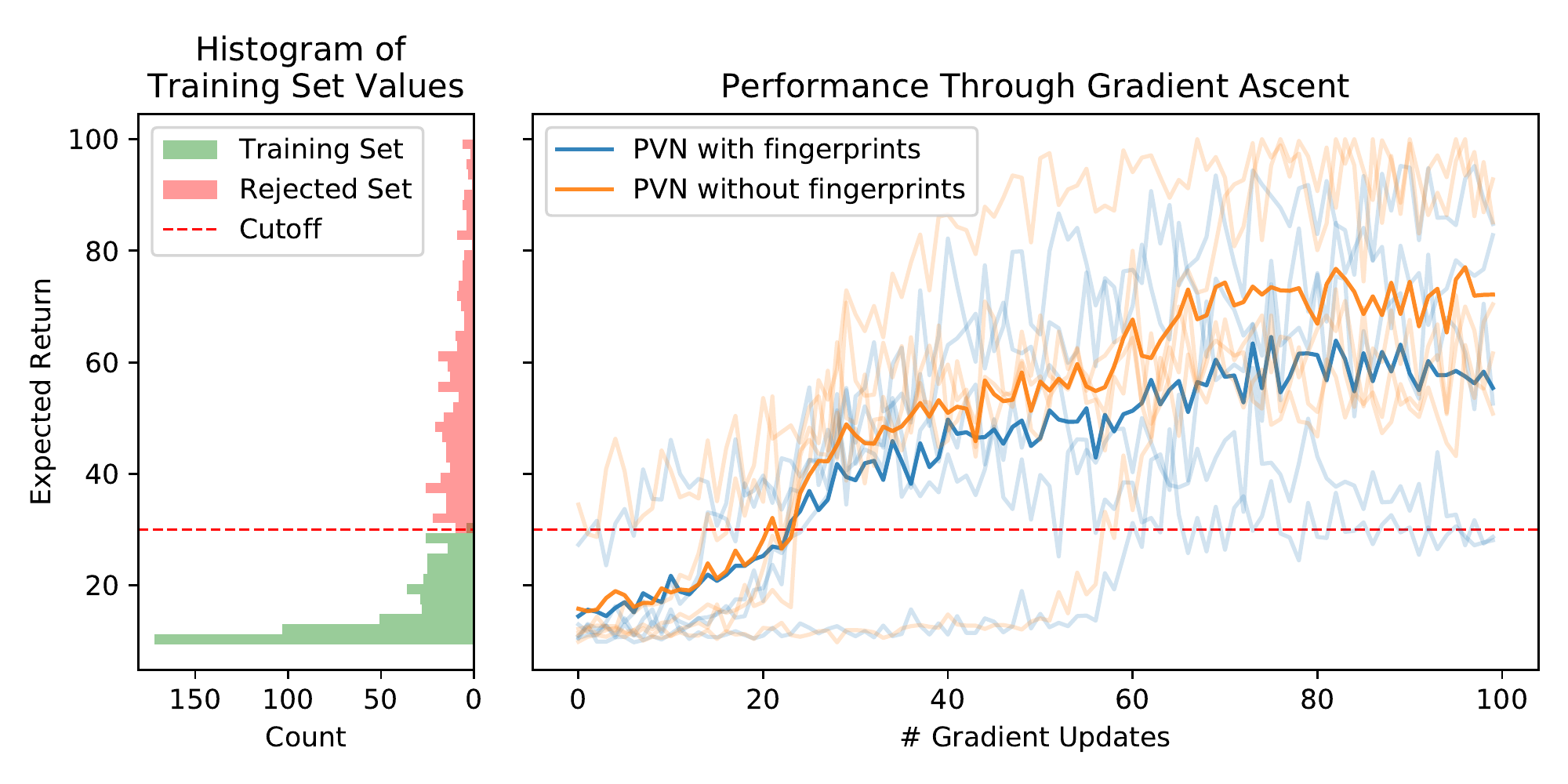}
  \caption{Linear Policy}
  \label{fig:cartpole_linear}
\end{subfigure}%
\hfill
\begin{subfigure}[t]{0.5\textwidth}
  \centering
  \includegraphics[scale=0.4]{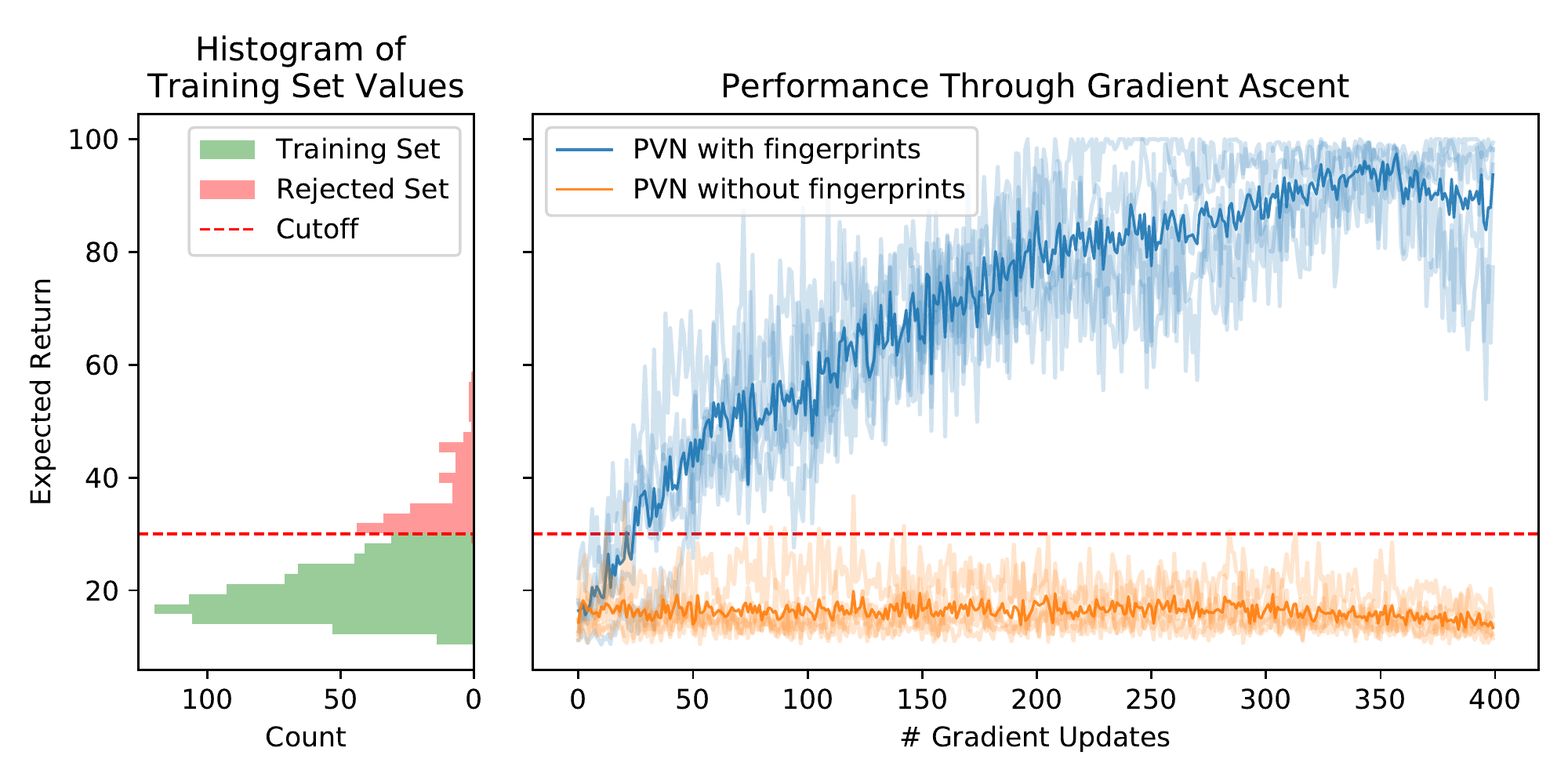}
  \caption{MLP Policy}
  \label{fig:cartpole_mlp}
\end{subfigure}
\end{center}
\caption{Plots showing histograms of training policies' expected returns and the performance of gradient ascent through a learned \pvn. The effects of Network Fingerprinting are drastic when using MLP policies.}
\label{fig:cartpoles}
\end{figure*}

\subsection{Cart Pole}
In the next set of experiments, we move to the Cart Pole environment with policy functions that map states to action distributions. The environment was run using 'CartPole-v0' in OpenAI Gym \cite{brockman2016openai}. This environment has an observation space of 4 features and 2 discrete actions, accelerating to the left and right. The main goal of this section is to show that Network Fingerprinting is crucial  to scaling \pvns to larger networks.

\textbf{Linear Policy.} First we start with the simplest case, a policy network consisting of a single linear layer with a softmax over the logits.

To generate a dataset in this setting, we start by creating a set of randomly generated linear policy networks, and getting a set of Monte-Carlo returns for each network. As the exact expected returns cannot be known, we instead have to approximate them from samples. Of course, with a large enough number of rollouts, the approximation can become accurate. This is now our dataset, where the policy networks are the inputs and the discretized distributions of sampled returns are the desired outputs to the \pvn. In these experiments, we generate a set of 1000 randomly initialized policies and run each policy for 100 episodes. The return distributions are discretized into 41 evenly sized bins.

When generating randomly initialized networks, some of the networks can have near optimal performance, making it difficult to show that our method can lead to policies better than the training set. In order to deal with this, we filter out of the training set any policy with an expected return above a specified level. This allows us to see if the \pvn\ generalizes outside of the distribution of policies seen in training and whether the gradient ascent can be effective and find substantially better policies. In our experiments, the training set consists of randomly initialized policies with an expected undiscounted return no higher than 30, while the maximum possible return is 100.

Once we have a trained \pvn, we then perform gradient ascent for 100 steps on 5 randomly sampled starting policies. When comparing methods, the same 5 policies are given to all. This allows for a fairer comparison, as one policy might be easier to improve than another.

The remaining figures  are designed in two parts. The left subplot is a histogram showing the distribution of expected returns achieved by the training set policies. In green is the training data used by the \pvn to train. In red is the data that was initially generated but then thrown away, as it was above our set expected return limit of 30. This allows us to show the performance achievable by gradient ascent relative to the training data. The right subplot is the performance of policies as they perform a number of gradient steps, in a lighter color. The darker line is the average policy ascent path. The starting policies are randomly selected by network initialization. Finally, the red dashed line marks the training set performance limit of 30.

We compare training \pvns in 2 ways: with and without Network Fingerprinting. As discussed in Section \ref{sec:probe}, performing network fingerprinting on linear networks is equivalent to flattening the network's weights and feeding them directly as input to a \pvn. As expected, both the flattened weights and network fingerprinting worked  similarly. Furthermore,  gradient ascent using either approach led to policies with an expected return of around 70, well above the training set limit of 30.

\textbf{MLP Policy.} The main challenge of building \pvns\ was to build a scalable mechanism allowing us to give a multi-layered policy network (MLP) as input to a \pvn. In these experiments we will show that network fingerprinting allows us to do so. 

We built the network fingerprinting with 20 probing states, leading to a policy fingerprint of 40 dimensions. These probing states were randomly initialized and trained jointly with the \pvn. We performed 400 steps of gradient ascent. The rest of the training procedure is exactly as described for the linear policy.

In Figure \ref{fig:cartpole_mlp}, we can clearly see that giving a flattened neural net as input to a \pvn\ does not work. On the other hand, network fingerprinting sees no scalability issues and allows the policy to improve up to the optimal policy. Furthermore, gradient ascent was consistently successful across the different starting policies, improving to near optimal performance from all starting points.

Comparing the histograms from Figures \ref{fig:cartpole_linear} and \ref{fig:cartpole_mlp}, we can notice that the distribution of generated linear policies has a long tail, with some randomly generated policies achieving near optimal performance. On the other hand, the distribution of randomly generated MLP policies has much lower ceiling of performance. As networks become larger, it becomes more difficult to randomly generate good policies.

\subsection{MuJoCo - Swimmer}
Our last set of experiments is on Swimmer, a MuJoCo~\cite{todorov2012mujoco} continuous control environment where the agent is a small worm that has to move forward by moving its joints. We test our approach on this task to show whether \pvns\ can scale to larger experiments. Also, as explained in Section \ref{sec:penfull}, state of the art RL algorithms tend to do poorly on Swimmer, compared to other MuJoCo tasks, because of the discounting. Agents optimizing for the discounted return learn to act in a myopic way, sacrificing long term gains, due to the discount. Since our approach can be used without discounting, we can avoid this problem by optimizing for the true objective directly.

In these experiments, we trained \pvns with Network Fingerprinting on a dataset of 2000 deterministic policies and 500 rollouts each to estimate their returns.  The return distributions are discretized into 51 evenly sized bins. Once the \pvn is trained, we do gradient ascent with 5 randomly initialized starting policies for 1000 steps each. The rest of the algorithmic details are in the Appendix. In Figure \ref{fig:Swimmer_mlp}, we compare the performance of the 5 policy ascents with 3 baselines, DDPG~\citep{lillicrap2015continuous}, SAC~\citep{haarnoja2018soft} and TD3~\citep{Fujimoto2018}, which are state of the art model-free RL algorithms. We can see that the set of policies all finished with an expected return above  all baselines. The best of the curves achieved expected returns around 250, substantially outperforming other algorithms.

\begin{figure}[]
  \centering
  \includegraphics[scale=0.4]{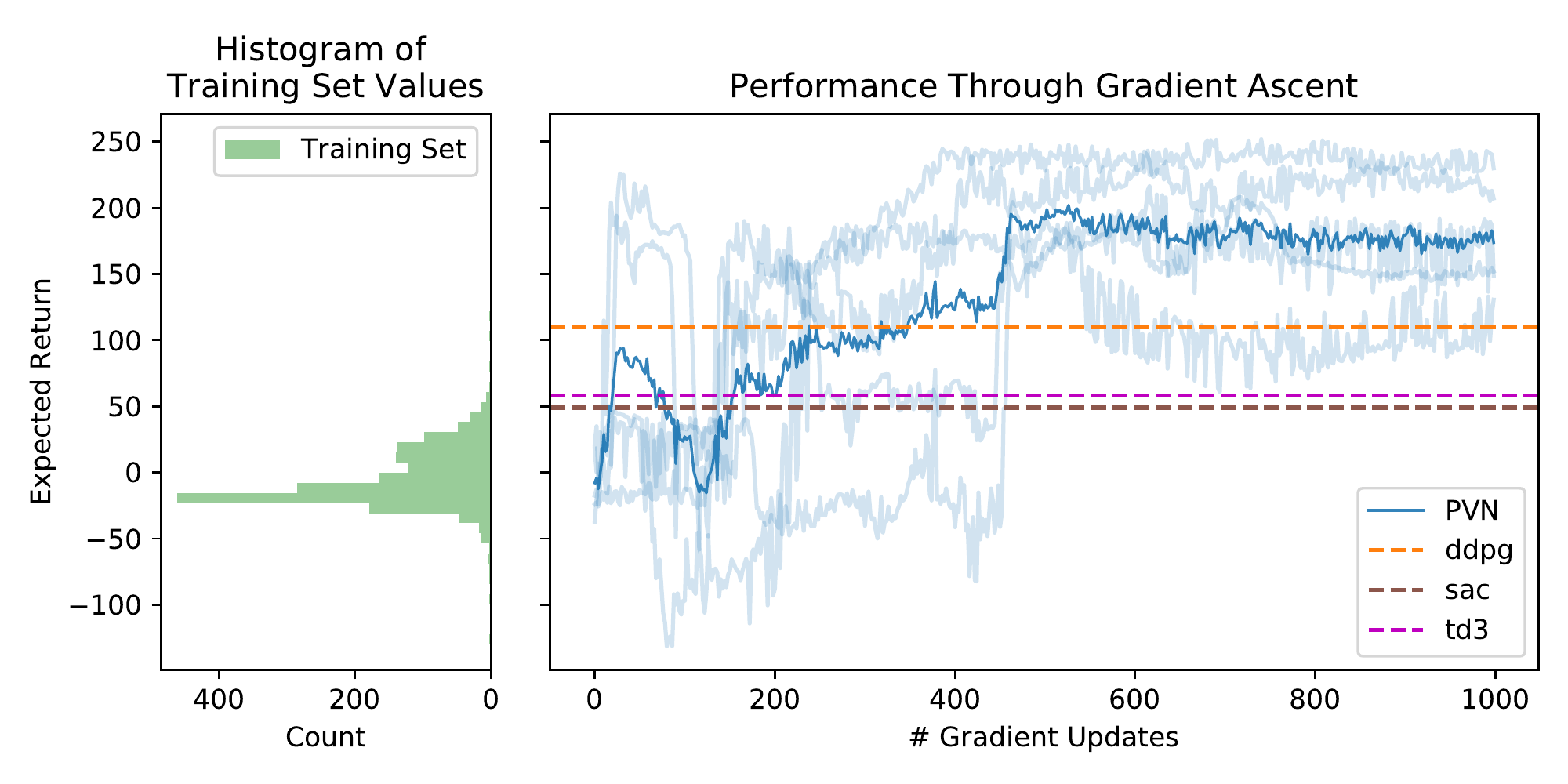}
  \caption{Gradient ascent performed on Swimmer. We compare the improvement of 5 starting policies and plot the average improvement in bold. Horizontal dashed lines are baselines. Their scores were taken from \href{https://spinningup.openai.com/en/latest/spinningup/bench.html}{https://spinningup.openai.com/en/latest/spinningup/bench.html}}
  \label{fig:Swimmer_mlp}
\end{figure}

\section{Related Work}

Methods that aim to solve RL problems by searching directly in policy space have a long history, and often different terminology. Sometimes they are characterized as black-box optimization, as they treat the mapping from policy parameters to return (or ``fitness'') as a black box, sometimes as evolutionary algorithms, with recent incarnations in~\citep{salimans2017evolution,such2017deep,mania2018simple}.
They are related to this work in two dimensions: a number of black-box methods also pursue a form of policy gradient ascent \citep{spall1992multivariate,peters2010relative,wierstra2014natural}, generally by employing a noisy form of finite differences. In theory, one could use finite differences to compute the exact gradient; however, there are too many parameters to make this a tractable solution. Our method on the other hand, is a low-variance but biased estimate of a policy's expected return, which in turn gives us a biased gradient. 
The second dimension of similarity is the analogue of our \pvns, also called `surrogate models' or `response surfaces' which aim to generalize across the fitness landscape, for example \citep{Booker1998,Box1951,Moore1996,ong2003evolutionary,loshchilov2012self}; see \citep{Jones2001} for an overview.
In contrast to our approach, which explicitly introduces an inductive bias to make suitable generalizations across policies, these methods make fewer assumptions and model only (often local) surface-level regularities of the fitness.

Our work has some similarities to synthetic gradients \cite{jaderberg2017decoupled}, which are networks that learn to output another network's gradients. However, our networks are never trained to output gradients; these are available to us as a byproduct of the architecture.

Generalized policy improvement \cite{barreto2017successor} finds a better policy as a mixture of a set of policies, which is a similar objective to ours, but the methodology is very different, as it relies on having Q-values for each policy. 

Universal Value Function Approximators \citep[UVFA,][]{schaul2015universal} are value functions that generalize across both states and goals. More specifically, since many policies can achieve the same goals, UVFAs output the value of the optimal policy for a certain goal. In contrast, our method generalizes across policies, regardless of their optimality, and is less complex because it does not depend on state.

Finally, our work can be considered a case of off-policy learning (\citet{sutton2018reinforcement}, \citet{precup2000eligibility}, \citet{munos2016safe}), a class of algorithms which allow one to evaluate and improve policies given data generated from a set of different policies. One major difference is that our method only looks at expected returns of policies, as opposed to all transitions generated by the set of policies, as is usually done.

\section{Conclusion and future work}
We introduced a network that can generalize in policy space, by taking policy fingerprints as inputs. These fingerprints are differentiable policy embeddings obtained by inspecting the policy's behaviour in a set of key states. We also described a novel policy gradient algorithm which is performed directly through the \pvnfull, allowing the computation of a policy gradient estimate for any policy, even if it has never been seen by the network.

\textbf{Extension to value functions.} Until now, we have only looked at \pvnfulls\ which output the expected return of a policy, from the initial state distribution. While this has benefits over the usual way of doing policy gradient, there are also disadvantages. Traditional RL algorithms can usually learn on-line, that is learn as more samples are seen, whereas our method requires entire trajectories before learning. This, however, does not have to be the case.
Our method is extendable to the state-dependent value function setting $V(s,\theta)$. This can give rise to zero-shot policy evaluation for a variety of algorithms. In actor-critic algorithms for instance, when the policy updates, the value function is lagging behind and requires samples from the new policy before it becomes accurate. Our method would allow a value function to generalize to unseen policies, meaning when the policy is updated, the value function would immediately update as well. This has the potential to improve data efficiency, as policies would not have to wait for value functions to catch up.

\textbf{Inductive Biases.} We designed \pvns\ in the simplest way possible, as feed-forward neural networks. However, the structure of the network is very important. Inductive biases incorporated into the architecture can substantially improve data efficiency and generalization in policy space \cite{wolpert1997no}. There is much structure in MDPs that can be leveraged.
Instead of simply building an MLP, one can build a state transition model and use this information to make value predictions. Other works, such as TreeQN \citep{farquhar2018treeqn}, the Predictron \citep{silver2017predictron} and Value Prediction Networks \cite{oh2017value} are examples of value functions built with inductive biases, looking to improve generalization.

\subsubsection*{Acknowledgments}
We'd like to acknowledge Simon Osindero, Kory Mathewson, Tyler Jackson, Chantal Remillard and Sasha Vezhnevets for useful discussions and feedback on the paper. Most importantly, we'd like to thank Emma Brunskill for inviting Jean Harb to her lab, where this work was started. Pierre-Luc Bacon is supported by the Facebook CIFAR AI chair program and IVADO. Doina Precup is a CIFAR fellow and is supported by the Canada CIFAR chair program.

\bibliographystyle{icml2020}
\bibliography{ms}

\begin{thebibliography}{45}
\providecommand{\natexlab}[1]{#1}
\providecommand{\url}[1]{\texttt{#1}}
\expandafter\ifx\csname urlstyle\endcsname\relax
  \providecommand{\doi}[1]{doi: #1}\else
  \providecommand{\doi}{doi: \begingroup \urlstyle{rm}\Url}\fi

\bibitem[Barreto et~al.(2017)Barreto, Dabney, Munos, Hunt, Schaul, van Hasselt,
  and Silver]{barreto2017successor}
Barreto, A., Dabney, W., Munos, R., Hunt, J.~J., Schaul, T., van Hasselt,
  H.~P., and Silver, D.
\newblock Successor features for transfer in reinforcement learning.
\newblock In \emph{Advances in neural information processing systems}, pp.\
  4055--4065, 2017.

\bibitem[Baxter \& Bartlett(2001)Baxter and Bartlett]{Baxter2001}
Baxter, J. and Bartlett, P.~L.
\newblock Infinite-horizon policy-gradient estimation.
\newblock \emph{J. Artif. Int. Res.}, 15\penalty0 (1):\penalty0 319–350,
  November 2001.
\newblock ISSN 1076-9757.

\bibitem[Bellemare et~al.(2017{\natexlab{a}})Bellemare, Dabney, and
  Munos]{Bellemare2017}
Bellemare, M.~G., Dabney, W., and Munos, R.
\newblock A distributional perspective on reinforcement learning.
\newblock In \emph{Proceedings of the 34th International Conference on Machine
  Learning, {ICML} 2017, Sydney, NSW, Australia, 6-11 August 2017}, pp.\
  449--458, 2017{\natexlab{a}}.

\bibitem[Bellemare et~al.(2017{\natexlab{b}})Bellemare, Dabney, and
  Munos]{bellemare2017distributional}
Bellemare, M.~G., Dabney, W., and Munos, R.
\newblock A distributional perspective on reinforcement learning.
\newblock In \emph{Proceedings of the 34th International Conference on Machine
  Learning-Volume 70}, pp.\  449--458. JMLR. org, 2017{\natexlab{b}}.

\bibitem[Booker et~al.(1998)Booker, Dennis, Frank, and Serafini]{Booker1998}
Booker, A., Dennis, J.~J., Frank, P., and Serafini, D.
\newblock Optimization using surrogate objectives on a helicopter test example.
\newblock \emph{Computational Methods in Optimal Design and Control}, 1998.

\bibitem[Box \& Wilson(1951)Box and Wilson]{Box1951}
Box, G. E.~P. and Wilson, K.~B.
\newblock On the experimental attainment of optimum conditions.
\newblock \emph{Journal of the Royal Statistical Society}, 13\penalty0
  (1):\penalty0 1--45, 1951.

\bibitem[Brockman et~al.(2016)Brockman, Cheung, Pettersson, Schneider,
  Schulman, Tang, and Zaremba]{brockman2016openai}
Brockman, G., Cheung, V., Pettersson, L., Schneider, J., Schulman, J., Tang,
  J., and Zaremba, W.
\newblock {OpenAI} gym.
\newblock \emph{arXiv preprint arXiv:1606.01540}, 2016.

\bibitem[Caruana(1997)]{Caruana1997}
Caruana, R.
\newblock Multitask learning.
\newblock \emph{Machine Learning}, 28\penalty0 (1):\penalty0 41--75, Jul 1997.

\bibitem[Cohn et~al.(1995)Cohn, Ghahramani, and Jordan]{Cohn1995}
Cohn, D.~A., Ghahramani, Z., and Jordan, M.~I.
\newblock Active learning with statistical models.
\newblock \emph{Journal of Artificial Intelligence Research}, 4:\penalty0
  129--145, 1995.

\bibitem[Dadashi et~al.(2019)Dadashi, Taiga, Roux, Schuurmans, and
  Bellemare]{dadashi2019value}
Dadashi, R., Taiga, A.~A., Roux, N.~L., Schuurmans, D., and Bellemare, M.~G.
\newblock The value function polytope in reinforcement learning.
\newblock \emph{arXiv preprint arXiv:1901.11524}, 2019.

\bibitem[Farquhar et~al.(2018)Farquhar, Rockt{\"a}schel, Igl, and
  Whiteson]{farquhar2018treeqn}
Farquhar, G., Rockt{\"a}schel, T., Igl, M., and Whiteson, S.
\newblock {TreeQN} and {ATreeC}: Differentiable tree planning for deep
  reinforcement learning.
\newblock In \emph{International Conference on Learning Representations}.
  International Conference on Learning Representations, 2018.

\bibitem[Fujimoto et~al.(2018)Fujimoto, van Hoof, and Meger]{Fujimoto2018}
Fujimoto, S., van Hoof, H., and Meger, D.
\newblock Addressing function approximation error in actor-critic methods.
\newblock In \emph{Proceedings of the 35th International Conference on Machine
  Learning, {ICML} 2018, Stockholmsm{\"{a}}ssan, Stockholm, Sweden, July 10-15,
  2018}, pp.\  1582--1591, 2018.

\bibitem[Glynn \& Olvera-Cravioto(2019)Glynn and Olvera-Cravioto]{Glynn2019}
Glynn, P.~W. and Olvera-Cravioto, M.
\newblock Likelihood ratio gradient estimation for steady-state parameters.
\newblock \emph{Stochastic Systems}, 9\penalty0 (2):\penalty0 83--100, June
  2019.

\bibitem[Haarnoja et~al.(2018)Haarnoja, Zhou, Abbeel, and
  Levine]{haarnoja2018soft}
Haarnoja, T., Zhou, A., Abbeel, P., and Levine, S.
\newblock Soft actor-critic: Off-policy maximum entropy deep reinforcement
  learning with a stochastic actor.
\newblock \emph{arXiv preprint arXiv:1801.01290}, 2018.

\bibitem[Jaderberg et~al.(2017{\natexlab{a}})Jaderberg, Czarnecki, Osindero,
  Vinyals, Graves, Silver, and Kavukcuoglu]{jaderberg2017decoupled}
Jaderberg, M., Czarnecki, W.~M., Osindero, S., Vinyals, O., Graves, A., Silver,
  D., and Kavukcuoglu, K.
\newblock Decoupled neural interfaces using synthetic gradients.
\newblock In \emph{Proceedings of the 34th International Conference on Machine
  Learning-Volume 70}, pp.\  1627--1635. JMLR. org, 2017{\natexlab{a}}.

\bibitem[Jaderberg et~al.(2017{\natexlab{b}})Jaderberg, Dalibard, Osindero,
  Czarnecki, Donahue, Razavi, Vinyals, Green, Dunning, Simonyan,
  et~al.]{jaderberg2017population}
Jaderberg, M., Dalibard, V., Osindero, S., Czarnecki, W.~M., Donahue, J.,
  Razavi, A., Vinyals, O., Green, T., Dunning, I., Simonyan, K., et~al.
\newblock Population based training of neural networks.
\newblock \emph{arXiv preprint arXiv:1711.09846}, 2017{\natexlab{b}}.

\bibitem[Jaderberg et~al.(2017{\natexlab{c}})Jaderberg, Mnih, Czarnecki,
  Schaul, Leibo, Silver, and Kavukcuoglu]{Jaderberg2017}
Jaderberg, M., Mnih, V., Czarnecki, W.~M., Schaul, T., Leibo, J.~Z., Silver,
  D., and Kavukcuoglu, K.
\newblock Reinforcement learning with unsupervised auxiliary tasks.
\newblock In \emph{5th International Conference on Learning Representations,
  {ICLR} 2017, Toulon, France, April 24-26, 2017, Conference Track
  Proceedings}, 2017{\natexlab{c}}.

\bibitem[Jones(2001)]{Jones2001}
Jones, D.~R.
\newblock A taxonomy of global optimization methods based on response surfaces.
\newblock \emph{Journal of Global Optimization}, 21:\penalty0 345--383, 2001.

\bibitem[Konda \& Tsitsiklis(2000)Konda and Tsitsiklis]{konda1999}
Konda, V.~R. and Tsitsiklis, J.~N.
\newblock Actor-critic algorithms.
\newblock In \emph{NIPS}, pp.\  1008--1014, 2000.

\bibitem[Lillicrap et~al.(2015)Lillicrap, Hunt, Pritzel, Heess, Erez, Tassa,
  Silver, and Wierstra]{lillicrap2015continuous}
Lillicrap, T.~P., Hunt, J.~J., Pritzel, A., Heess, N., Erez, T., Tassa, Y.,
  Silver, D., and Wierstra, D.
\newblock Continuous control with deep reinforcement learning.
\newblock \emph{arXiv preprint arXiv:1509.02971}, 2015.

\bibitem[Loshchilov et~al.(2012)Loshchilov, Schoenauer, and
  Sebag]{loshchilov2012self}
Loshchilov, I., Schoenauer, M., and Sebag, M.
\newblock Self-adaptive surrogate-assisted covariance matrix adaptation
  evolution strategy.
\newblock In \emph{Proceedings of the 14th annual conference on Genetic and
  evolutionary computation}, pp.\  321--328, 2012.

\bibitem[Lyle et~al.(2019)Lyle, Bellemare, and Castro]{lyle2019comparative}
Lyle, C., Bellemare, M.~G., and Castro, P.~S.
\newblock A comparative analysis of expected and distributional reinforcement
  learning.
\newblock In \emph{Proceedings of the AAAI Conference on Artificial
  Intelligence}, volume~33, pp.\  4504--4511, 2019.

\bibitem[Mania et~al.(2018)Mania, Guy, and Recht]{mania2018simple}
Mania, H., Guy, A., and Recht, B.
\newblock Simple random search provides a competitive approach to reinforcement
  learning.
\newblock \emph{arXiv preprint arXiv:1803.07055}, 2018.

\bibitem[Moore \& Schneider(1996)Moore and Schneider]{Moore1996}
Moore, A.~W. and Schneider, J.
\newblock Memory-based stochastic optimization.
\newblock In \emph{Advances in Neural Information Processing Systems}, 1996.

\bibitem[Munos et~al.(2016)Munos, Stepleton, Harutyunyan, and
  Bellemare]{munos2016safe}
Munos, R., Stepleton, T., Harutyunyan, A., and Bellemare, M.
\newblock Safe and efficient off-policy reinforcement learning.
\newblock In \emph{Advances in Neural Information Processing Systems}, pp.\
  1054--1062, 2016.

\bibitem[Oh et~al.(2017)Oh, Singh, and Lee]{oh2017value}
Oh, J., Singh, S., and Lee, H.
\newblock Value prediction network.
\newblock In \emph{Advances in Neural Information Processing Systems}, pp.\
  6118--6128, 2017.

\bibitem[Ong et~al.(2003)Ong, Nair, and Keane]{ong2003evolutionary}
Ong, Y.~S., Nair, P.~B., and Keane, A.~J.
\newblock Evolutionary optimization of computationally expensive problems via
  surrogate modeling.
\newblock \emph{AIAA journal}, 41\penalty0 (4):\penalty0 687--696, 2003.

\bibitem[Peters et~al.(2010)Peters, Mulling, and Altun]{peters2010relative}
Peters, J., Mulling, K., and Altun, Y.
\newblock Relative entropy policy search.
\newblock In \emph{Twenty-Fourth AAAI Conference on Artificial Intelligence},
  2010.

\bibitem[Precup(2000)]{precup2000eligibility}
Precup, D.
\newblock Eligibility traces for off-policy policy evaluation.
\newblock \emph{Computer Science Department Faculty Publication Series}, pp.\
  ~80, 2000.

\bibitem[Salimans et~al.(2017)Salimans, Ho, Chen, Sidor, and
  Sutskever]{salimans2017evolution}
Salimans, T., Ho, J., Chen, X., Sidor, S., and Sutskever, I.
\newblock Evolution strategies as a scalable alternative to reinforcement
  learning.
\newblock \emph{arXiv preprint arXiv:1703.03864}, 2017.

\bibitem[Schaul et~al.(2015)Schaul, Horgan, Gregor, and
  Silver]{schaul2015universal}
Schaul, T., Horgan, D., Gregor, K., and Silver, D.
\newblock Universal value function approximators.
\newblock In \emph{International Conference on Machine Learning}, pp.\
  1312--1320, 2015.

\bibitem[Schulman et~al.(2016)Schulman, Moritz, Levine, Jordan, and
  Abbeel]{schulman2016}
Schulman, J., Moritz, P., Levine, S., Jordan, M.~I., and Abbeel, P.
\newblock High-dimensional continuous control using generalized advantage
  estimation.
\newblock In \emph{International Conference on Learning Representations
  (ICLR)}, 2016.

\bibitem[Silver et~al.(2017)Silver, van Hasselt, Hessel, Schaul, Guez, Harley,
  Dulac-Arnold, Reichert, Rabinowitz, Barreto, et~al.]{silver2017predictron}
Silver, D., van Hasselt, H., Hessel, M., Schaul, T., Guez, A., Harley, T.,
  Dulac-Arnold, G., Reichert, D., Rabinowitz, N., Barreto, A., et~al.
\newblock The predictron: End-to-end learning and planning.
\newblock In \emph{Proceedings of the 34th International Conference on Machine
  Learning-Volume 70}, pp.\  3191--3199. JMLR. org, 2017.

\bibitem[Spall et~al.(1992)]{spall1992multivariate}
Spall, J.~C. et~al.
\newblock Multivariate stochastic approximation using a simultaneous
  perturbation gradient approximation.
\newblock \emph{IEEE transactions on automatic control}, 37\penalty0
  (3):\penalty0 332--341, 1992.

\bibitem[Such et~al.(2017)Such, Madhavan, Conti, Lehman, Stanley, and
  Clune]{such2017deep}
Such, F.~P., Madhavan, V., Conti, E., Lehman, J., Stanley, K.~O., and Clune, J.
\newblock Deep neuroevolution: Genetic algorithms are a competitive alternative
  for training deep neural networks for reinforcement learning.
\newblock \emph{arXiv preprint arXiv:1712.06567}, 2017.

\bibitem[Sutton(1984)]{sutton1984}
Sutton, R.~S.
\newblock \emph{Temporal credit assignment in reinforcement learning}.
\newblock PhD thesis, University of Massachusetts Amherst, 1984.

\bibitem[Sutton \& Barto(2018)Sutton and Barto]{sutton2018reinforcement}
Sutton, R.~S. and Barto, A.~G.
\newblock \emph{Reinforcement learning: An introduction}.
\newblock MIT press, 2018.

\bibitem[Sutton et~al.(2000)Sutton, McAllester, Singh, and
  Mansour]{sutton2000policy}
Sutton, R.~S., McAllester, D.~A., Singh, S.~P., and Mansour, Y.
\newblock Policy gradient methods for reinforcement learning with function
  approximation.
\newblock In \emph{Advances in neural information processing systems}, pp.\
  1057--1063, 2000.

\bibitem[Sutton et~al.(2011)Sutton, Modayil, Delp, Degris, Pilarski, White, and
  Precup]{Sutton2011}
Sutton, R.~S., Modayil, J., Delp, M., Degris, T., Pilarski, P.~M., White, A.,
  and Precup, D.
\newblock Horde: a scalable real-time architecture for learning knowledge from
  unsupervised sensorimotor interaction.
\newblock In \emph{10th International Conference on Autonomous Agents and
  Multiagent Systems {(AAMAS} 2011), Taipei, Taiwan, May 2-6, 2011, Volume
  1-3}, pp.\  761--768, 2011.

\bibitem[Thomas(2014)]{thomas2014}
Thomas, P.
\newblock Bias in natural actor-critic algorithms.
\newblock In \emph{Proceedings of the 31th International Conference on Machine
  Learning, {ICML} 2014, Beijing, China, 21-26 June 2014}, pp.\  441--448,
  2014.

\bibitem[Todorov et~al.(2012)Todorov, Erez, and Tassa]{todorov2012mujoco}
Todorov, E., Erez, T., and Tassa, Y.
\newblock Mujoco: A physics engine for model-based control.
\newblock In \emph{2012 IEEE/RSJ International Conference on Intelligent Robots
  and Systems}, pp.\  5026--5033. IEEE, 2012.

\bibitem[van~den Oord et~al.(2016)van~den Oord, Kalchbrenner, and
  Kavukcuoglu]{Oord2016}
van~den Oord, A., Kalchbrenner, N., and Kavukcuoglu, K.
\newblock Pixel recurrent neural networks.
\newblock In \emph{Proceedings of the 33nd International Conference on Machine
  Learning, {ICML} 2016, New York City, NY, USA, June 19-24, 2016}, pp.\
  1747--1756, 2016.

\bibitem[Wierstra et~al.(2014)Wierstra, Schaul, Glasmachers, Sun, Peters, and
  Schmidhuber]{wierstra2014natural}
Wierstra, D., Schaul, T., Glasmachers, T., Sun, Y., Peters, J., and
  Schmidhuber, J.
\newblock Natural evolution strategies.
\newblock \emph{The Journal of Machine Learning Research}, 15\penalty0
  (1):\penalty0 949--980, 2014.

\bibitem[Wolpert \& Macready(1997)Wolpert and Macready]{wolpert1997no}
Wolpert, D.~H. and Macready, W.~G.
\newblock No free lunch theorems for optimization.
\newblock \emph{IEEE transactions on evolutionary computation}, 1\penalty0
  (1):\penalty0 67--82, 1997.

\bibitem[{Xi-Ren Cao} \& {Yat-Wah Wan}(1998){Xi-Ren Cao} and {Yat-Wah
  Wan}]{Cao1998}
{Xi-Ren Cao} and {Yat-Wah Wan}.
\newblock Algorithms for sensitivity analysis of markov systems through
  potentials and perturbation realization.
\newblock \emph{IEEE Transactions on Control Systems Technology}, 6\penalty0
  (4):\penalty0 482--494, July 1998.
\newblock ISSN 2374-0159.

\end{thebibliography}

\clearpage

\onecolumn
\icmltitle{Supplementary Material}
\appendix
\counterwithin{figure}{section}
\counterwithin{table}{section}

\section{Polytope Experiment Details}
\label{app:polytope}
\subsection{Markov Decision Process Specifications}
To describe the 2 state MDP used in our polytope experiments, we use the following convention~\citep{dadashi2019value}:
$$r(s_i, a_j) = \hat r[i \times |\mathcal{A}| + j]$$
$$P(s_k|s_i, a_j) = \hat P [i \times |\mathcal{A}| + j][k]$$
where our MDP has the following properties.
$$|\mathcal{A}| = 2, \gamma = 0.8$$
$$\hat r = [-0.45, -0.1, 0.5, 0.5]$$
$$\hat P = [[0.6, 0.4], [0.99, 0.01], [0.2, 0.8], [0.99, 0.01]]$$

\subsection{Visualization of Predictions}
\begin{figure}[h]
\centering
\begin{subfigure}[t]{.4\textwidth}
  \centering
  \includegraphics[width=.95\linewidth]{figs/polytope/training_polytope_no_corners.pdf}
  \caption{Exact policy values}
  \label{fig:training_polytope2}
\end{subfigure}%
\begin{subfigure}[t]{.4\textwidth}
  \centering
  \includegraphics[width=.95\linewidth]{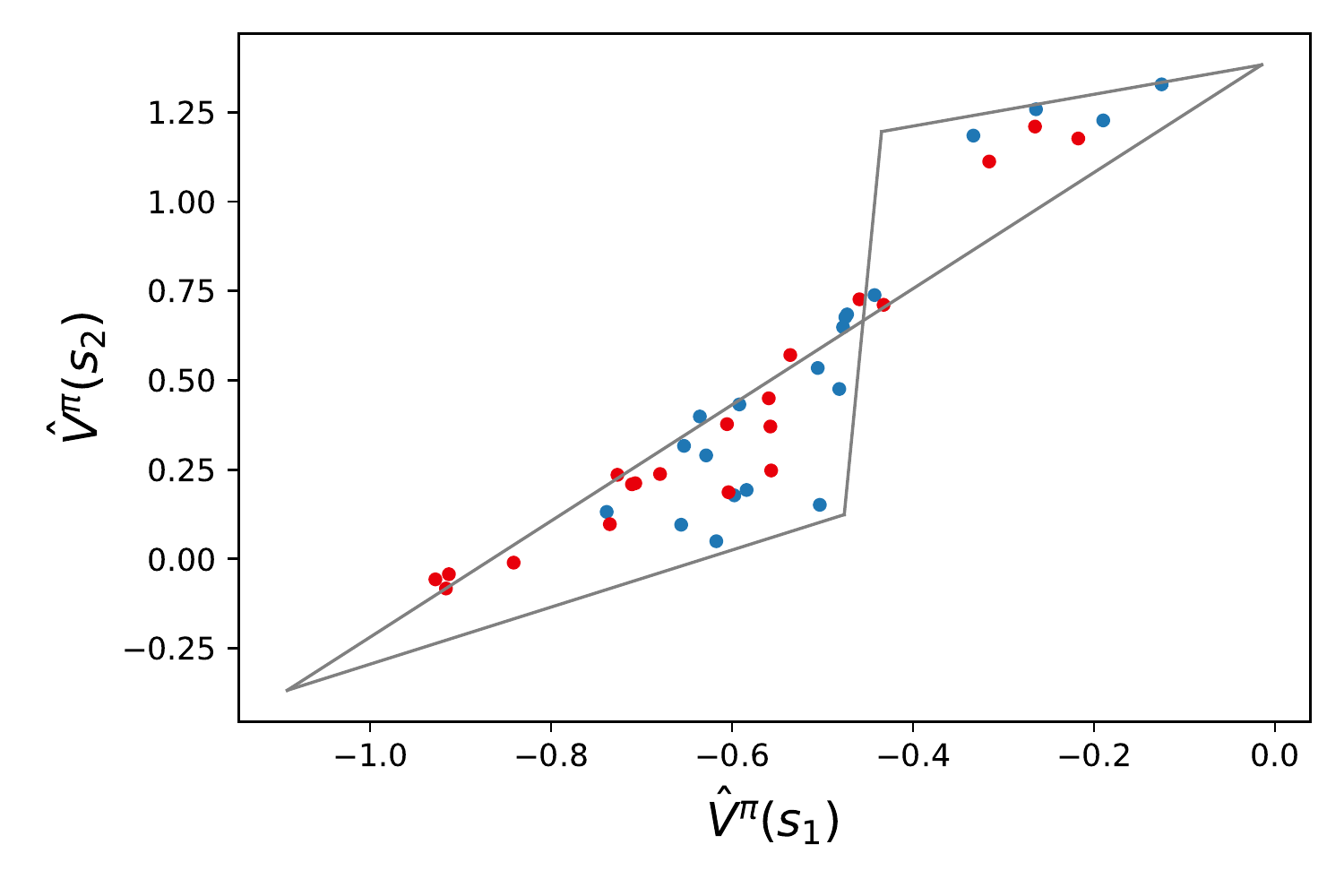}
  \caption{Policy values predicted by the trained PVN}
  \label{fig:trained_polytope2}
\end{subfigure}
\caption{Results of training a network to predict policy values, projected in a value polytope. The training set is in blue and the test set is in red.}
\label{fig:polytope_data2}
\end{figure}

In the polytope experiments, 40 points were randomly sampled and split into equally sized training and test sets of 20 points. In Figure \ref{fig:training_polytope2}, we show the sampled points projected in the value polytope. In contrast, Figure \ref{fig:trained_polytope2} shows the values of the points predicted by a trained \pvn.

\clearpage
\section{Hyperparameters}

\begin{table}[h]
    \centering
\caption{Hyperparameters used in various experiments. The last three rows are only applicable when using probing states.}
\vskip 0.15in
 \begin{tabular}{l r r r r}
 \toprule
 Parameter & Polytope & Cartpole Linear & Cartpole MLP & Swimmer\\ [0.5ex]
 \midrule
 Policy architecture (hidden layer sizes) &  N/A &[] & [30] & [30]\\
 PVN Architecture (hidden layer sizes) &     [50] & [80] & [80] & [80]\\
 NN activations &                       ReLU & ReLU & ReLU & ReLU\\
 Temperature &                          N/A & 3 & 3 & 3\\
 \# Bins &                              N/A & 41 & 41 & 51\\
 Grad ascent learning rate &            0.1 & 0.001 & 0.001 & 0.002\\
 Grad ascent optimizer &                SGD & Adam & Adam & Adam\\
 Grad ascent steps &                    100 & 100 & 400 & 1000\\
 Discount factor ($\gamma$) &           0.8 & 1 & 1 & 1\\
 Batch size &                           32 & 32 & 32 & 32\\
 PVN learning rate &                    0.01 & 0.003 & 0.003 & 0.003\\
 PVN optimizer &                        RMSProp & Adam & Adam & Adam\\
 Training steps &                       20000 & 3000 & 3000 & 5000\\
 \# Policies &                          20 & 1000 & 1000 & 2000\\
 \# Returns per policy &                N/A & 100 & 100 & 500\\
 Training set performance limit &       N/A & 30 & 30 & N/A \\
 \midrule\midrule
 Train probing states &                 N/A & True & True & True\\
 Randomly generate probing states &     N/A & True & True & True\\
 \# Probing states &                    N/A & 20 & 20 & 20\\
 \bottomrule
\end{tabular}
\end{table}

\section{Dataset collection algorithm}
\begin{algorithm}
   \caption{Collect dataset of policies for \pvn\ training}
   \label{alg:pen}
\begin{algorithmic}
    \STATE Choose number of policies $K$ and rollouts $B$
    \STATE $\mathcal{D} \leftarrow \varnothing$

    \FOR{policy $i=1$ {\bfseries to} $K$}
    \STATE Initialize policy $\pi_i$ parameters $\theta_i$ using Glorot Initialization
    \STATE $R \leftarrow \varnothing$

    \FOR{rollout $b=1$ {\bfseries to} $B$}
    \STATE Sample Monte-Carlo return $G_{MC}$ using policy $\pi_{\theta_i}$
    \STATE $R \leftarrow R \cup G_{MC}$
    \ENDFOR
    \STATE $\mathcal{D} \leftarrow \mathcal{D} \cup (\theta_i, R)$
    \ENDFOR
    \STATE \textbf{return} $\mathcal{D}$
\end{algorithmic}
\end{algorithm}

\end{document}